\documentclass[runningheads]{llncs}
\usepackage[T1]{fontenc}
\usepackage{graphicx}
\usepackage{graphicx}
\usepackage{amsmath,amssymb}
\usepackage{booktabs}
\usepackage{multirow}
\usepackage{subcaption}
\usepackage{xcolor}
\usepackage{xspace}
\usepackage{tikz}
\usepackage{float}

\usepackage{hyphenat}

\usetikzlibrary{arrows.meta,positioning,fit,backgrounds,calc,shapes.geometric}
\usepackage[hidelinks]{hyperref}
\usepackage{orcidlink}

\newcommand{\method}{\textsc{Scalp}\xspace} 
\begin{document}
\title{SCALP: Semi-Supervised Statistical Shape Modeling from
       Imperfect 3D Photogrammetry via Landmark-Anchored Spectral Warping}
\titlerunning{SCALP}
%

\author{Nawazish Khan\inst{1} \and
Sanjay Bhandari\inst{1, 2} \and
Sarang Joshi\inst{1, 3} \and
Alzbeta Novotna\inst{4} \and
Tiffany Jeong\inst{4} \and
Loretta Bowman\inst{4} \and
Michael Hernandez\inst{4} \and
Tobi Somorin\inst{4} \and
Viraj Govani\inst{4} \and
Jesse Goldstein\inst{4} \and
Shireen Elhabian\inst{1,2}}

\authorrunning{N. Khan et al.}

\institute{Scientific Computing and Imaging Institute, University of Utah, Salt Lake City, UT 84112, USA \\
\email{\{nawazish.khan, shireen\}@sci.utah.edu}\and
Kahlert School of Computing, University of Utah, Salt Lake City, UT 84112, USA \and
Department of Biomedical Engineering, University of Utah, Salt Lake City, UT 84112, USA \and
Division of Pediatric Plastic Surgery, UPMC Children's Hospital of Pittsburgh, Pittsburgh, PA 15213, USA
}
\maketitle              
\begin{abstract}
Correspondence-based statistical shape modeling (SSM) is vital for population-level morphometric analysis, but conventional pipelines assume clean, fully registered surfaces. Real-world clinical photogrammetry scans are often noisy, partial, and cluttered, hindering the adoption of radiation-free surface imaging as a safe alternative to computed tomography (CT) for infant craniosynostosis. We present \method{} (\textbf{S}emi-supervised \textbf{C}orrespondence via l\textbf{A}ndmark \textbf{L}ocalization and s\textbf{P}ectral warping), a two-stage framework that constructs consistent shape models directly from raw, imperfect surface scans. First, a semi-supervised Point Transformer leverages a small expert-annotated dataset alongside a large unlabeled cohort to accurately localize craniofacial landmarks with minimal annotation overhead. Second, these landmarks anchor a Laplace--Beltrami spectral deformation of an anatomical template, generating dense correspondences while naturally isolating the cranium from peripheral scanning clutter without manual preprocessing. Experiments on infant photogrammetry scans demonstrate that \method{} consistently outperforms state-of-the-art unsupervised point-cloud approaches, offering a clinically practical pathway toward objective, radiation-free head shape analysis.

\end{abstract}

%
%
%
\section{Introduction}
\label{sec:intro}

Statistical shape modeling (SSM) is a cornerstone of computational anatomy, enabling the quantitative analysis of anatomical variability across populations. By establishing dense correspondence between individual shapes, SSM provides compact, anatomically meaningful representations that support disease characterization, surgical planning, treatment evaluation, and the discovery of population-level morphological patterns~\cite{shapeworks,deepssm,khan2023statistical}. However, correspondence-based SSM fundamentally relies on high-quality geometric input: complete, clean, registered surfaces with consistent topology~\cite{shapeworks,mesh2ssm}. These assumptions are rarely satisfied in routine clinical imaging, where surface scans acquired from low-cost, non-invasive modalities are often noisy, incomplete, unregistered, and contaminated by anatomy outside the region of interest. Constructing reliable statistical shape models directly from such imperfect surface data remains an open challenge and a major barrier to translating accessible imaging modalities into quantitative clinical practice.

Craniosynostosis provides a critical setting in which to address this challenge. Affecting approximately one in 2,500 infants, the premature fusion of one or more cranial sutures disrupts normal cranial growth, potentially leading to elevated intracranial pressure and neurodevelopmental impairment~\cite{craniosynostosis_overview}. To move beyond subjective visual evaluation, clinical severity scoring pipelines, such as CranioRate, utilize SSMs to place a patient on an objective, quantitative deformity continuum~\cite{bhalodia2020,craniorate,mendoza2014}. However, existing workflows depend heavily on computed tomography (CT). The associated ionizing radiation, high cost, and limited repeatability of CT severely restrict its utility for longitudinal monitoring. More importantly, clinicians cannot repeatedly expose an infant to radiation over the first two years of life simply to track rapid, ongoing changes in cranial morphology. 

Three-dimensional photogrammetry offers an attractive alternative, providing a radiation-free, fast, and repeatable modality that can be deployed at any routine clinic visit~\cite{meulstee2017,rodriguezflorez2017,abdel2023reliability,abdel2023sagittal}. Recent clinical evidence confirms that severity scores derived from 3D photography can closely match CT-derived scores~\cite{bruce20233d}. However, establishing these metrics has historically required extensive, manual, expert-driven preprocessing and rigid alignment of the photogrammetric scans before any correspondence could be computed~\cite{bruce20233d}. This reliance on intensive manual curation creates a major bottleneck, rendering existing pipelines unscalable across multi-institution clinical consortia. Furthermore, raw photogrammetric data exemplifies the imperfect surface data that derails automated SSM pipelines: it captures only external topography (including skin, hair, and clothing) rather than isolated cranial bones, and contains significant non-anatomical clutter such as the neck and shoulders.

These limitations expose a fundamental trade-off in current learning-based correspondence methods. Supervised landmark localization provides the strict anatomical grounding required to isolate the cranium but demands prohibitive annotation datasets and exhibits limited generalization across differing acquisition systems. Conversely, unsupervised point-cloud correspondence frameworks eliminate manual labeling entirely but lack anatomical awareness; when confronted with raw, unsegmented photogrammetric scans, they erroneously distribute correspondence points across peripheral structures like the neck and shoulders. 

\begin{figure}[t]
\vspace{-10pt}
\centering
\includegraphics[width=\textwidth]{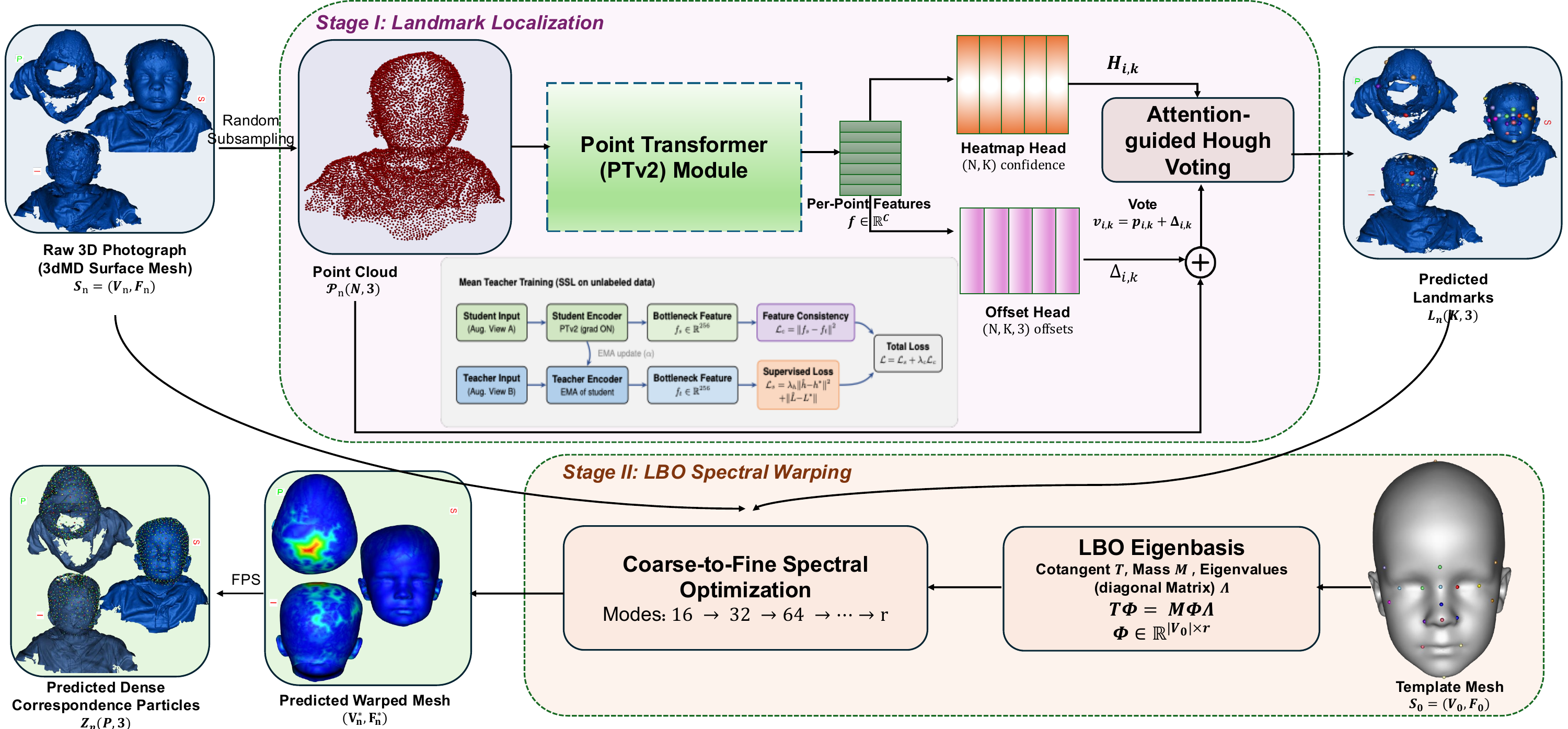}
\caption{Overview of the \method{} framework. Stage I (Landmark Localization): A raw photogrammetric mesh $S_n$ is downsampled via FPS to point cloud $\mathcal{P}_n$. A Point Transformer (PTv2) backbone extracts features for parallel Heatmap and Offset heads, localized into landmarks $L_n$ via attention-guided Hough voting. Training uses a semi-supervised Mean Teacher paradigm with a feature consistency loss on unlabeled data. Stage II (LBO Spectral Warping): An anatomical template mesh $S_0$ undergoes coarse-to-fine deformation restricted to the first $r$ low-frequency eigenfunctions $\Phi$ of its Laplace--Beltrami Operator (LBO). This warp is anchored by the predicted landmarks $L_n$, outputting a topologically consistent mesh and dense cranial correspondence particles $Z_n$ that naturally filter out peripheral scanning clutter.}
\label{fig:pipeline}
\vspace{-20pt}
\end{figure}

To overcome these challenges, we present \textbf{S}emi-supervised \textbf{C}orrespondence via l\textbf{A}ndmark \textbf{L}ocalization and s\textbf{P}ectral warping (\method{}), a two-stage framework that automates dense, anatomically consistent SSM construction directly from raw, unprocessed photogrammetric scans (Fig.~\ref{fig:pipeline}). In Stage I, \method{} breaks the annotation-generalization bottleneck by learning robust craniofacial landmarks through a geometry-aware semi-supervised framework, achieving reliable anatomical grounding using a minimal annotation budget easily provided by a single clinical fellow. In Stage II, these predicted landmarks guide a template-fitting deformation parameterized within a band-limited Laplace--Beltrami spectral basis. Unlike traditional registration methods, such as Thin Plate Splines (TPS) or non-rigid ICP, which enforce surface smoothness via soft energy penalties that can violate anatomical constraints under data pressure, our framework restricts deformation to the lowest $r$ eigenmodes. This band-limiting establishes a hard geometric constraint that prevents high-frequency folding entirely by construction, making the pipeline inherently robust to the missing data, sampling artifacts, and peripheral clutter characteristic of routine surface scans.

The main contributions of this work are:
\begin{itemize}

    \item \textit{Annotation-Efficient Landmark Localization:} We introduce a novel architectural pairing for raw photogrammetric point clouds, combining a Point Transformer v2 (PTv2) backbone with a Hough voting head. By integrating confidence-filtered pseudo-labeling with a Mean Teacher consistency objective tailored specifically to geometric point-cloud augmentations, our framework achieves robust landmark localization using only 76 expert-annotated subjects while maintaining strong generalization across domain shifts.
    
    \item \textit{Hard-Constrained Spectral Template Warp:} We propose a landmark-guided Laplace--Beltrami spectral deformation engine for establishing dense anatomical correspondence. In contrast to prior spectral works utilizing the operator strictly for local descriptors or generative shape synthesis, we deploy it for subject-specific template fitting. By optimizing directly within the low-frequency eigenbasis, we eliminate high-frequency mesh folding by construction, providing a mathematical robustness guarantee against missing regions and severe outlier clutter.
    
    \item \textit{An End-to-End Scalable Clinical Pipeline:} We deliver an automated pipeline that transforms raw clinical photogrammetry scans into dense correspondence particles suitable for downstream morphometric analysis. This framework eliminates manual preprocessing and explicit segmentation, closing the technical capability gap preventing scalable, radiation-free, multi-institution longitudinal assessment of craniofacial deformities.

\end{itemize}

\section{Related Work}
\label{sec:related}

\noindent\textit{Statistical shape modeling.}
Statistical shape modeling (SSM) relies on establishing anatomically consistent point-to-point correspondences across a population. Establishing this correspondence automatically has a long lineage, spanning from optimization-based minimum-description-length (MDL) formulations and spherical harmonic parameterizations~\cite{davies2002minimum,styner2006} to the entropy-minimizing particle systems of ShapeWorks~\cite{shapeworks}. While these classical methods remain highly effective on segmented anatomical surfaces, they share fundamental limitations: they assume complete, clean, and consistently connected meshes, optimize the entire cohort jointly at considerable computational cost, and assume linear shape variation. Consequently, these assumptions break down when applied to raw photogrammetry scans, which are intrinsically noisy, incomplete, unregistered, and heavily dominated by non-anatomical structures such as the neck, shoulders, and clothing.

\noindent\textit{Learning- and Template-Based Correspondence.} Deep learning has shifted correspondence estimation from iterative optimization toward direct representation learning. Among these, image-based approaches like DeepSSM regress shape descriptors directly from 2D/3D images but remain fully supervised and reliant on pre-optimized reference shape models~\cite{deepssm}. To operate directly on raw geometries, unsupervised and self-supervised point-cloud frameworks, including Point2SSM/++~\cite{point2ssm,point2ssmpp}, DG-AE~\cite{achlioptas2018learning}, DPC~\cite{dpc}, and SC3K~\cite{sc3k}, predict correspondences from unordered points. However, they implicitly assume all observed inputs belong to the target anatomy, erroneously distributing correspondence across peripheral scanner clutter (e.g., neck and shoulders, artifacts). Alternatively, template-based architectures like Mesh2SSM/++~\cite{mesh2ssm,mesh2ssmpp} establish consistent topologies by deforming a common reference mesh, but they typically require clean, segmented CT or MRI surfaces and lack the landmark guidance needed to handle the severe geometric misalignment or missing data common in raw photogrammetry.

\noindent\textit{Photogrammetry for craniosynostosis.}
Three-dimensional photogrammetry has emerged as an attractive, radiation-free imaging modality for craniofacial assessment~\cite{meulstee2017,rodriguezflorez2017}. Early clinical applications established that photogrammetry-based severity measurements can replicate traditional CT-based metrics, but required intensive manual landmark annotation, mesh decimation, and alignment before establishing correspondence, limiting multi-center utility~\cite{bruce20233d}. Subsequent deep learning works focused primarily on binary diagnostic classification rather than dense morphological modeling~\cite{dejong2020,schaufelberger2022}, or used scans for automated regional head measurements without learning underlying population-level dense correspondences~\cite{abdel2023reliability,abdel2023sagittal}. The closest frameworks remain partial solutions: Schaufelberger \emph{et al.}~\cite{schaufelberger2022} developed a public radiation-free craniosynostosis model tailored for classification, while Elkhill \emph{et al.}~\cite{elkhill2023} and Cruz Guerrero \emph{et al.}~\cite{cruzguerrero2024} proposed automated landmark localization and registration pipelines. However, these latter methods depend heavily on scanner-specific appearance cues and fail to transfer robustly to texture-free geometry from alternative acquisition hardware (as shown in Sec.~\ref{sec:quant}). Our approach bridges this gap, enabling an automated, radiation-free alternative to traditional CT-based severity metrics, such as CranioRate scores~\cite{bhalodia2020,craniorate,mendoza2014}, directly from raw scans with minimal supervision.


\noindent\textit{Landmark Localization on Point Clouds.} Processing native point clouds avoids the discretization artifacts of voxelization~\cite{maturana2015} and the geometric loss of multi-view projections~\cite{su2015}. Evolving from PointNet's permutation-invariant global pooling~\cite{pointnet}, hierarchical and graph-based models like PointNet++ and DGCNN successfully capture localized geometric context~\cite{pointnetpp,dgcnn}. Point transformers represent the natural continuation of this trajectory, leveraging self-attention to adaptively model long-range spatial relationships over irregular surface topologies~\cite{pct,pointtransformer,ptv2}. While intrinsic geometric networks offer appealing pose invariance~\cite{masci2015}, they typically require clean meshes with consistent connectivity and pre-computed shape operators, making them brittle when applied to noisy, partial, and variably-tessellated photogrammetry scans.

\noindent\textit{Spectral Shape Analysis.} The Laplace--Beltrami operator (LBO) provides an intrinsic, ordered eigenbasis where low frequencies capture global structure and high frequencies encode local detail~\cite{levy2006,reuter2006}, underpinning classic spectral descriptors~\cite{sun2009hks,aubry2011wks} and functional maps for non-rigid matching~\cite{ovsjanikov2012,litany2017,donati2020}. Truncating this spectral subspace provides inherent band-limiting regularization, yielding remarkable robustness to noise, sampling artifacts, and connectivity variations. Within surface tracking and modeling paradigms, spectral architectures differ fundamentally: some focus on generative synthesis from a learned prior (e.g., ToothForge~\cite{toothforge}), while others regularize surface deformations directly within the eigenbasis. Compared to classical spatial registration techniques like Coherent Point Drift (CPD), Bayesian CPD, and non-rigid ICP~\cite{cpd,bcpd,amberg2007}, which enforce smoothness via soft penalties that routinely collapse under outlier noise or missing regions, band-limited spectral bases offer a hard geometric constraint that inherently prevents high-frequency surface folding during severe partiality~\cite{rodola2017partial} and large inter-subject variations~\cite{dai2018non}.
\section{Method}
\label{sec:method}

\subsection{Problem Formulation}
\label{sec:setup}
We establish dense anatomical correspondence across a cohort of raw 3D craniofacial photogrammetry scans $\{\mathcal{S}_i\}_{i=1}^{M}$, where each subject is represented by an unstructured triangular mesh $\mathcal{S}_i=(\mathbf{V}_i,\mathbf{F}_i)$. These meshes exhibit high variability in vertex count and connectivity, and contain artifacts, missing data, and non-cranial structures (e.g., neck and shoulders). While clinical stereophotogrammetry natively outputs meshes, our framework explicitly requires this representation to perform Laplace--Beltrami spectral deformation via the cotangent discretization of the mesh Laplacian. Our objective is to estimate an ordered set of $P$ correspondence particles $\mathbf{Z}_i=\{\mathbf{z}_i^{1},\ldots,\mathbf{z}_i^{P}\}\subset \mathbb{R}^{3}$ per subject, where the $p$-th particle represents the identical anatomical location across the cohort to enable downstream statistical analyses. 
Direct estimation of dense correspondence from raw photogrammetry is challenging because the input surfaces are noisy, incomplete, inconsistently aligned, and contain large regions that are irrelevant to (Fig.~\ref{fig:pipeline}). First, a semi-supervised landmark localization network predicts a sparse set of $K$ clinically meaningful craniofacial anchors directly from the raw scan. Second, these landmarks constrain a band-limited Laplace--Beltrami spectral deformation of a common template mesh, producing anatomically consistent dense correspondence particles while intrinsically filtering out non-cranial peripheral clutter.

\subsection{Stage I: Semi-supervised landmark localization}
\label{sec:stage1}

The first stage of \method{} predicts a sparse set of clinically defined craniofacial landmarks directly from an imperfect photogrammetry scan. Accurate landmark localization is challenging because the input consists of unordered point clouds without texture, while expert annotations are available for only a small subset of the training data. To address these challenges, we combine a Point Transformer backbone with a Hough voting localization head and a semi-supervised training strategy that leverages both labeled and unlabeled scans.

\noindent\textbf{Point Transformer backbone.} Each input scan is uniformly sampled to $N$ points,
$
\mathcal{P}=\{\mathbf{p}_1,\ldots,\mathbf{p}_N\},
\mathbf{p}_i\in\mathbb{R}^{3}$, 
and processed by Point Transformer v2 network, which learns hierarchical point features while preserving the irregular geometric structure of the surface~\cite{ptv2}. Unlike voxel- or image-based representations, point transformers operate directly on unordered point sets, avoiding discretization artifacts and naturally accommodating varying point densities. The network outputs a feature vector $\mathbf{f}_i\in\mathbb{R}^{C}$ for every input point, capturing both local surface geometry and long-range anatomical context.

\noindent\textbf{Hough voting for landmark prediction.} Rather than directly regressing landmark coordinates, we formulate localization as a Hough voting problem \cite{qi2019votenet}.
A heatmap head emits a per-point confidence map
$\mathbf{H}\in[0,1]^{N\times K}$, supervised toward a Gaussian of bandwidth $\sigma$
centred on each true landmark $\boldsymbol{\ell}^{*}_k$,
\begin{equation}
  H^{*}_{ik}=\exp\!\Bigl(-\|\mathbf{p}_i-\boldsymbol{\ell}^{*}_k\|^2/2\sigma^2\Bigr),
  \label{eq:heatmap_gt}
\end{equation}
while an offset head predicts a displacement $\boldsymbol{\Delta}_{ik}$ from each
point to each landmark, turning every point into a vote
$\mathbf{v}_{ik}=\mathbf{p}_i+\boldsymbol{\Delta}_{ik}$. The landmark is the
consensus of those votes, weighted by the heatmap through a sharpening temperature
$\tau$,
\begin{equation}
  \hat{\boldsymbol{\ell}}_k=\sum_{i=1}^{N} w_{ik}\,\mathbf{v}_{ik},
  \qquad w_{ik}=\frac{\exp(H_{ik}/\tau)}{\sum_{j}\exp(H_{jk}/\tau)},
  \label{eq:voting}
\end{equation}
which, the votes being unconstrained, may lie anywhere in $\mathbb{R}^3$ and so
escapes the quantisation floor. The loss for the both the heads are given by
\begin{equation}
  \mathcal{L}_{offset}
   = \frac{1}{K}\sum_{k=1}^{K}\bigl\|\hat{\boldsymbol{\ell}}_k-\boldsymbol{\ell}^{*}_k\bigr\|_2,
   \qquad \mathcal{L}_{conf} = \frac{1}{NK}\sum_{i=1}^{N}\sum_{k=1}^{K}\bigl(H_{ik}-H^{*}_{ik}\bigr)^2,
  \label{eq:sup_loss}
\end{equation}






\noindent\textbf{Semi-supervised learning.} Manual landmark annotation is expensive and limits the scalability of supervised learning. We close this gap with confidence-gated self-training. We therefore train the landmark detector using a semi-supervised Mean Teacher framework, combining a small labeled dataset $
\mathcal{D}_L
=
\{(\mathcal{P}_i,\mathbf{L}_i)\}$ with a substantially larger unlabeled dataset $
\mathcal{D}_U
=
\{\mathcal{P}_j\}$. The student network is optimized using standard supervised losses on labeled examples, consisting of an offset regression loss coming from the offset head and a voting confidence loss for heatmap head,

\begin{equation}
\mathcal{L}_{sup}
=
\mathcal{L}_{offset}
+
\lambda_c
\mathcal{L}_{conf}.
\end{equation}
where, $\lambda_c$ balances the learning between the two heads of the network.





For unlabeled scans, we add a Mean Teacher~\cite{tarvainen2017meanteacher}
consistency term: a weight-averaged teacher
$\theta'_t=\alpha\theta'_{t-1}+(1-\alpha)\theta_t$ and the student each process a
different augmentation of every unlabeled scan, and we match their bottleneck
features
\begin{equation}
  \mathcal{L}_{\mathrm{cons}}=\bigl\|\mathbf{z}_s-\operatorname{sg}(\mathbf{z}_t)\bigr\|_2^2
  \label{eq:cons_loss}
\end{equation}
rather than their landmark predictions ($\mathbf{z}_s,\mathbf{z}_t$ are the student
and teacher bottleneck features and $\operatorname{sg}$ is stop-gradient). This setting of matching features is a deliberate choice as the unlabeled landmark predictions are still noisy they
would only propagate their own error, whereas the latent geometry is stable from the
outset. The overall training objective is

\begin{equation}
\mathcal{L}
=
\mathcal{L}_{sup}
+
\lambda_u
\mathcal{L}_{cons},
\end{equation}

where $\lambda_u$ balances supervised and unsupervised learning.

By combining transformer-based geometric reasoning with semi-supervised learning, the landmark detector learns robust anatomical representations from only a limited number of expert annotations while effectively exploiting the large collection of unlabeled photogrammetry scans available in clinical practice.

\subsection{Stage II: Landmark-guided spectral template warping}
\label{sec:stage2}
Given the localized landmarks from Stage I, the second stage establishes dense, population-wide shape correspondence by morphing a common anatomical template onto each subject scan. To fulfill our objective of a completely CT-free pipeline across both training and inference, we adopt a normative infant head surface template obtained from the open-source \texttt{CraniumPy} toolbox~\cite{abdelalim2023craniumpy,abdel2023reliability}, which is built upon the public statistical shape model described by Schaufelberger \emph{et al.}~\cite{schaufelberger2022}. This template is derived entirely from 3D stereophotogrammetric surface scans, rather than CT imaging, of normocephalic infants, and covers the full cranial vault and face with a standardized resolution of $|\mathbf{V}_0| = 20{,}000$ vertices. The $K = 23$ landmark vertex locations ($\mathbf{v}_{m_k}$) are predefined and manually pre-mapped onto fixed vertex indices of this template mesh by an expert clinical observer to act as anchor points. Because the template is delivered as a clean, fully connected triangulated mesh, it allows for the precise computation of cotangent weights and eigenfunctions directly from its fixed connectivity, completely bypassing the topological noise or lack of mesh connectivity present in the raw target scans. By deforming the template inside its own intrinsic, band-limited basis, we embed geometric smoothness directly into the optimization search space instead of enforcing it post-hoc. Let $\boldsymbol{\Phi}\in\mathbb{R}^{|\mathbf{V}_0|\times r}$ hold the first $r$ non-trivial eigenvectors of the template's cotangent Laplace--Beltrami operator, ordered by eigenvalue $\lambda_1\le\cdots\le\lambda_r$; these represent the template's natural vibration modes, ranging from global structural variations to fine local details~\cite{levy2006}. The number of spectral modes $r$ dictates the expressiveness of the deformation space; we present the heuristic and ablation experiment in Sect. \ref{sec:experiments} to help selecting it, thereby optimizing the trade-off between geometric flexibility and high-frequency noise suppression. Parameterizing the deformation by spectral coefficients $\mathbf{W}\in\mathbb{R}^{r\times3}$,
\begin{equation}
  \mathbf{V}_{\mathrm{def}}(\mathbf{W})=\mathbf{V}_0+\boldsymbol{\Phi}\mathbf{W},
  \label{eq:spectral_deformation}
\end{equation}
confines the warp strictly to the first $r$ harmonics and rules out high-frequency folding outright. The coefficients $\mathbf{W}$ are then recovered by minimizing an objective function consisting of a surface-fit, a smoothness, and a landmark alignment term,
\begin{equation}
  \mathbf{W}^{\star}=\arg\min_{\mathbf{W}}\;
  \mathcal{L}_{\mathrm{data}}(\mathbf{W})
  + \lambda_{\mathrm{reg}}\,\mathcal{L}_{\mathrm{reg}}(\mathbf{W})
  + \beta_{\mathrm{lm}}\,\mathcal{L}_{\mathrm{lm}}(\mathbf{W}).
  \label{eq:stage2_obj}
\end{equation}
The data term is a two-way Chamfer distance between the deformed template
$\mathbf{v}_j=\mathbf{V}_{\mathrm{def}}(\mathbf{W})_j$ and the subject
$\tilde{\mathbf{Y}}$,
\begin{equation}
  \mathcal{L}_{\mathrm{data}}
   = \sum_{j} w_j \min_{k}\|\mathbf{v}_j-\tilde{\mathbf{y}}_k\|_2^2
   + \beta_{\mathrm{2w}}\sum_{k}\min_{j}\|\tilde{\mathbf{y}}_k-\mathbf{v}_j\|_2^2 ,
  \label{eq:chamfer}
\end{equation}
whose reverse direction (activated midway through optimization) stops the template
collapsing onto a subset of the surface.
The smoothness term penalizes each coefficient row $\mathbf{w}_j$ by its eigenvalue,
\begin{equation}
  \mathcal{L}_{\mathrm{reg}}=\sum_{j=1}^{r}\lambda_j\,\|\mathbf{w}_j\|_2^2 ,
  \label{eq:reg}
\end{equation}
biasing the fit toward the low-frequency modes that carry gross cranial shape, and
the landmark term ties the template's landmark vertices $\mathbf{v}_{m_k}$ to the
predictions,
\begin{equation}
  \mathcal{L}_{\mathrm{lm}}=\sum_{k=1}^{K}\bigl\|\mathbf{v}_{m_k}-\hat{\boldsymbol{\ell}}_k\bigr\|_2^2 ,
  \label{eq:landmark_loss}
\end{equation}
with $\beta_{\mathrm{lm}}$ annealed so the Chamfer term sets a coarse fit before the landmarks tighten the face. We optimize coarse-to-fine, opening the active modes from a small low-frequency set up to all $r$ over several stages so that global shape settles before detail. Because every converged mesh shares the template's connectivity, a single farthest-point sampling of the template fixes the $P$ correspondence particles on it, thereby dense, anatomy-confined, and obtained vertex indices are preserved across the population, yielding dense one-to-one anatomical correspondence  $\{\mathbf{Z}_i\}$ across the population without requiring any pairwise registration or alignment.
\section{Experiments}
\label{sec:experiments}
\noindent\textbf{Data and Splits.} All data consist of texture-free, deidentified infant craniofacial surface meshes acquired via 3dMD stereophotogrammetry from the CranioRate consortium~\cite{craniorate}, spanning normative subjects and patients with various craniofacial conditions including metopic craniosynostosis and cleft palate. Scans are natively calibrated in physical millimeters ($\sim$66k--160k vertices); consequently, all geometric errors throughout our evaluation are reported directly in millimeters. All experiments utilize a fixed cohort of 217 subjects, partitioned into 204 for training/SSM construction and 13 held-out test subjects. To ensure a representative test set, we partition the cohort using K-means clustering over global shape embeddings from a pretrained PointM2AE~\cite{zhang2022point} and uniformly sample from each cluster; test subjects are strictly excluded from all training, model selection, and pseudo-labeling. Annotation supervision is intentionally sparse: an expert clinician annotated 23 landmarks for only 76 subjects. Using the same clustering approach to maximize morphological diversity, these annotated subjects are divided into 50 training, 13 validation, and 13 test subjects. The landmark localization network is trained using these 50 annotated training subjects alongside the remaining unlabeled consortium scans via our semi-supervised paradigm (Section~\ref{sec:stage1}).

\subsection{Baselines and Comparison Setup}
\label{sec:baselines}
We compare \method{} against the current state of the art in learning dense correspondence from point clouds. Our primary baseline is Point2SSM++~\cite{point2ssmpp}, together with all unsupervised correspondence methods included in its benchmark: Point2SSM~\cite{point2ssm}, DGCNN-based autoencoder (DG-AE)~\cite{achlioptas2018learning}, Intrinsic Structural Representation (ISR)~\cite{isr}, 
Deep Point Correspondence (DPC)~\cite{dpc}, and SC3K~\cite{sc3k}. Optimization-based particle shape modeling (PSM) methods such as ShapeWorks~\cite{shapeworks} are not included because they require complete, segmented, and pre-aligned surfaces, assumptions that are violated by raw photogrammetry scans. We likewise exclude ToothForge~\cite{toothforge}, whose objective is spectral shape generation rather than subject-specific correspondence estimation. We also evaluate zero-shot landmark localization performance of Stage I against DGCNN-based craniofacial landmark detector model of Elkhill \emph{et al.}~\cite{elkhill2023}, representing the closest published method for automatic landmark prediction from craniofacial surface scans.

\subsection{Evaluation Metrics}
\label{sec:metrics}
We evaluate landmark localization and dense correspondence quality using complementary geometric and statistical metrics.
\noindent\textit{Geometric \& Structural Fidelity:} Stage I landmark accuracy is quantified via Mean Radial Error (MRE) and Success Detection Rate (SDR). For baseline correspondence methods lacking explicit landmark predictions, we report Landmark Localization Error (LLE) to measure anatomical fidelity~\cite{shapeworks}. Let $\mathbf{G}_i=\{\mathbf{g}_i^{(k)}\}_{k=1}^{K}$ be ground-truth landmarks for subject $i$, and $\mathbf{P}_i=\{\mathbf{p}_i^{(j)}\}_{j=1}^{N}$ be predicted correspondence particles. LLE computes the average distance from each ground-truth landmark to its nearest particle across all $N$ subjects and $K$ landmarks: $\mathrm{LLE} = \frac{1}{NK} \sum_{i,k} \min_j \|\mathbf{g}_i^{(k)} - \mathbf{p}_i^{(j)}\|_2$. Surface reconstruction from these particles is further validated using point-to-surface (P2S) and warp-based surface-to-surface (Warp S2S) distances.
\noindent\textit{Statistical Shape Model Quality:} Following standard protocols~\cite{shapeworks}, the resulting SSMs are evaluated using compactness, generalization (CD), and specificity (CD) to quantify statistical efficiency, reconstruction capacity, and anatomical realism, respectively. Each metric is evaluated as a function of retained principal modes and summarized as the Area Under the Curve (AUC), normalized by the maximum number of modes.

\subsection{Evaluation Protocol}
\label{sec:protocol}
All methods are evaluated under an identical protocol on a test set of $N=13$ subjects with 23 expert-annotated landmarks, where each method predicts 1,024 particles per subject. To ensure a fair comparison against unsupervised baselines that model the entire input scan (including neck/shoulders), all geometric and statistical metrics are restricted to a common craniofacial region of interest (ROI) defined by a template bounding box. Reconstructed shapes are aligned to the template via Generalized Procrustes Analysis (GPA) before applying this ROI filter; for baselines, an equivalent bounding box filter is applied after initial ICP alignment during preprocessing to isolate cranial anatomy. Conversely, all landmark-based evaluations are performed directly in each subject's native coordinate frame. Baseline particles are inverse-transformed using their estimated ICP transformations before computing distances to manual annotations, while \method{} particles are transformed analogously, ensuring localization errors are measured directly in the native geometry.

\subsection{Quantitative Results}
\label{sec:quant}
We ask two questions: whether the landmark transformer module localizes landmarks accurately enough to
anchor the warp, and whether the full pipeline yields dense correspondence particles competitive with the existing methods.
\begin{table}[t]
\vspace{-20pt}
\centering
\caption{Landmark localization accuracy on 13 held-out test subjects. We evaluate our full semi-supervised Stage I framework against its purely supervised ablation (primary comparison) and the external model of Elkhill \emph{et al.}~\cite{elkhill2023} applied zero-shot without retraining (secondary cross-domain evaluation). MRE: Mean Radial Error; SDR@$t$: Success Detection Rate within a $t$\,mm error threshold. Bold values indicate the top-performing method.}
\label{tab:stage-1-metrics}
\setlength{\tabcolsep}{5pt}
\resizebox{\textwidth}{!}{
\begin{tabular}{lcccc}
\toprule
Method & MRE (mm) & Median (mm) & SDR@4 & SDR@10 \\
\midrule
Elkhill et al.~\cite{elkhill2023} (zero-shot) & $65.66 \pm 19.35$ & $67.17$ & $4.2\%$ & $10.4\%$ \\
Stage I (supervised training) & $13.46 \pm 11.94$ & $11.06$ & $12.7\%$ & $49.2\%$ \\
\textbf{Ours Stage I (semi-supervised)} & $\mathbf{8.48 \pm 5.42}$ & $\mathbf{7.17}$ & $\mathbf{18.1\%}$ & $\mathbf{69.6\%}$ \\
\bottomrule
\end{tabular}%
}
\vspace{-20pt}
\end{table}
We first evaluate the landmark localization accuracy of Stage I to determine whether it provides a sufficiently stable anchor for the downstream template deformation pipeline. To isolate the explicit performance gains of our semi-supervised framework, we establish a primary baseline by training a purely supervised counterpart of the same Point Transformer v2 architecture, restricting its training strictly to the limited labeled cohort. As reported in Table~\ref{tab:stage-1-metrics},
the full \method{} framework substantially outperforms this matched supervised-only baseline. This improvement directly validates the efficacy of our semi-supervised strategy, demonstrating that the integration of a Mean Teacher consistency objective with confidence-filtered pseudo-labeling successfully leverages the unlabeled cohort to capture robust geometric priors that a small labeled dataset alone cannot resolve. As a secondary finding to analyze robustness under domain shift, we also compare our method against the closest public craniofacial landmark detector by Elkhill \emph{et al.}~\cite{elkhill2023}. This DGCNN-based architecture was trained on a massive external cohort of craniofacial 3D photographs and is deployed here zero-shot using its released checkpoint without retraining. The results reveal a pronounced performance degradation for the model of Elkhill \emph{et al.}~\cite{elkhill2023} on our data distribution. This gap highlights a critical limitation in the generalization of texture-dependent landmark detectors when confronted with significant variations in acquisition hardware, noise characteristics, and raw geometry. In contrast, \method{} maintains precise localization by learning directly from native, texture-free point cloud structures. Crucially, the level of accuracy achieved by \method{} is highly sufficient to reliably initialize the landmark-anchored spectral warping in Stage II, effectively neutralizing the risk of downstream correspondence instability or topological folding due to catastrophic landmark mislocalization.
\begin{table}[htbp]
\vspace{-20pt}
\centering
\caption{Quantitative results. Best result per metric is \textbf{bold}. CD = Chamfer Distance, P2S = Point-to-Surface, Warp S2S = Warp-based Surface-to-Surface, LLE = Landmark Localization Error.}
\label{tab:results_final}
\resizebox{\textwidth}{!}{%
\begin{tabular}{lrrrrrrr}
\toprule
\textbf{Metric} & \textbf{DG-AE} & \textbf{DPC} & \textbf{ISR} & \textbf{Point2SSM} & \textbf{Point2SSM++} & \textbf{SC3K} & \textbf{SCALP} \\
\midrule
\multicolumn{8}{l}{\textit{Surface \& Correspondence Metrics}} \\
\midrule
P2S (mm) $\downarrow$ & $18.661 \pm 11.049$ & $13.348 \pm 8.469$ & $19.734 \pm 10.670$ & $18.610 \pm 11.282$ & $18.136 \pm 10.813$ & $25.896 \pm 7.660$ & $\mathbf{1.478 \pm 0.497}$ \\
Warp S2S (mm) $\downarrow$ & $9.220 \pm 6.559$ & $4.054 \pm 2.831$ & $4.387 \pm 1.700$ & $5.050 \pm 3.720$ & $4.181 \pm 2.974$ & $8.368 \pm 2.840$ & $\mathbf{1.456 \pm 0.492}$ \\
LLE (mm) $\downarrow$ & $8.087 \pm 1.812$ & $5.549 \pm 0.705$ & $7.129 \pm 0.950$ & $6.139 \pm 0.808$ & $6.364 \pm 0.642$ & $9.374 \pm 1.118$ & $\mathbf{3.520 \pm 0.542}$ \\
\midrule
\multicolumn{8}{l}{\textit{SSM Metrics}} \\
\midrule
Compactness $\uparrow$ & $0.937$ & $0.935$ & $0.925$ & $0.926$ & $0.928$ & $0.945$ & $\mathbf{0.990}$ \\
Generalization (CD) $\downarrow$ & $55.780$ & $103.577$ & $110.761$ & $122.601$ & $112.095$ & $58.847$ & $\mathbf{1.112}$ \\
Specificity (CD) $\downarrow$ & $135.370$ & $163.842$ & $143.863$ & $153.728$ & $137.794$ & $84.861$ & $\mathbf{9.604}$ \\
\bottomrule
\end{tabular}
}
\vspace{-20pt}
\end{table}

\begin{figure}
  \vspace{-10pt}
  \centering
  
  \begin{minipage}[t]{0.48\textwidth}
    \centering
    \includegraphics[width=\textwidth]{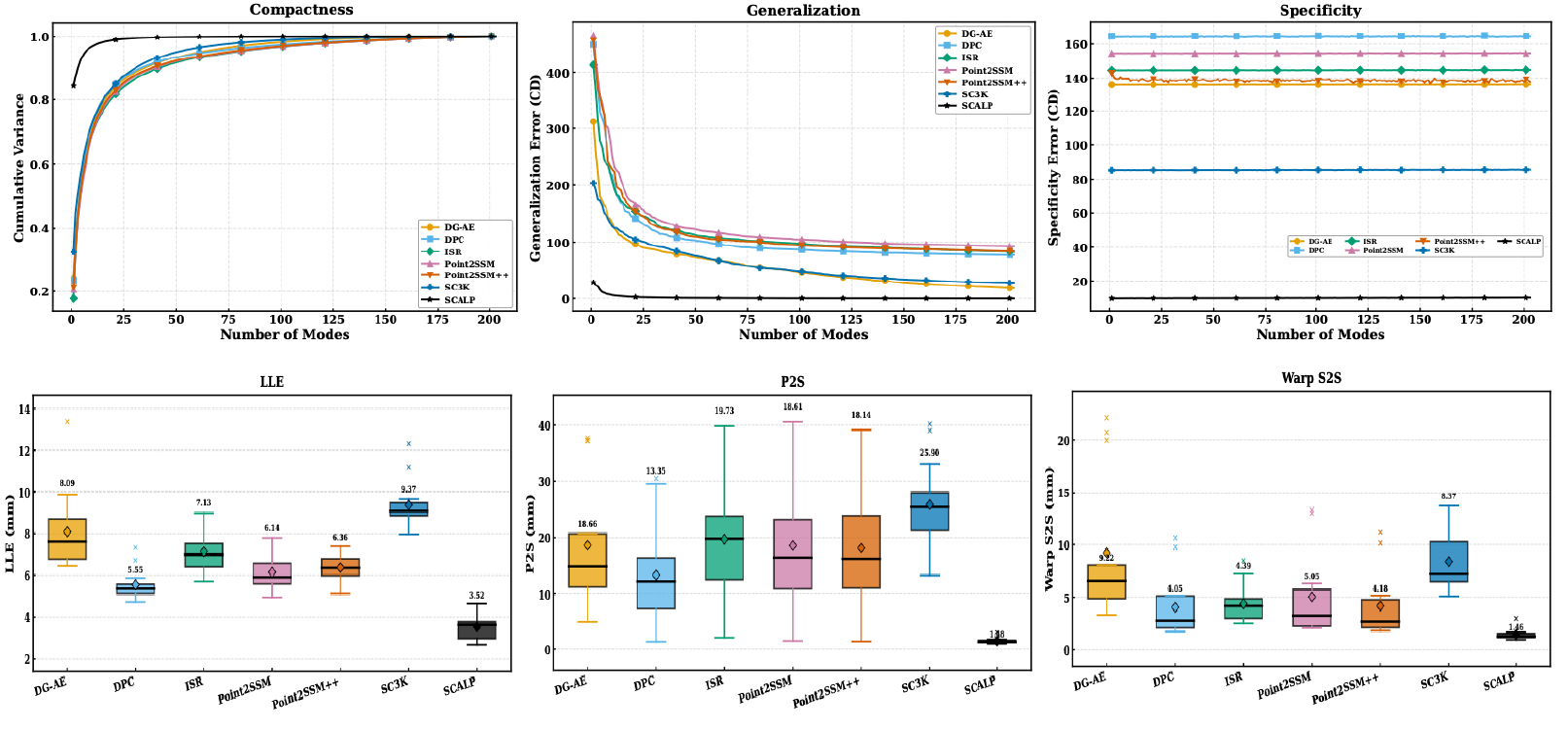}
    \caption{Quantitative evaluation against baselines. Top: Statistical Shape Model (SSM) quality metrics (Compactness, Generalization, and Specificity) plotted across principal modes. Bottom: Box plots showing test cohort distributions for geometric errors: LLE, P2S, and Warp S2S metrics.}
    
    \label{fig-ssm-quant}
  \end{minipage}
  \hfill 
  \begin{minipage}[t]{0.48\textwidth}
    \centering
    \includegraphics[width=\textwidth]{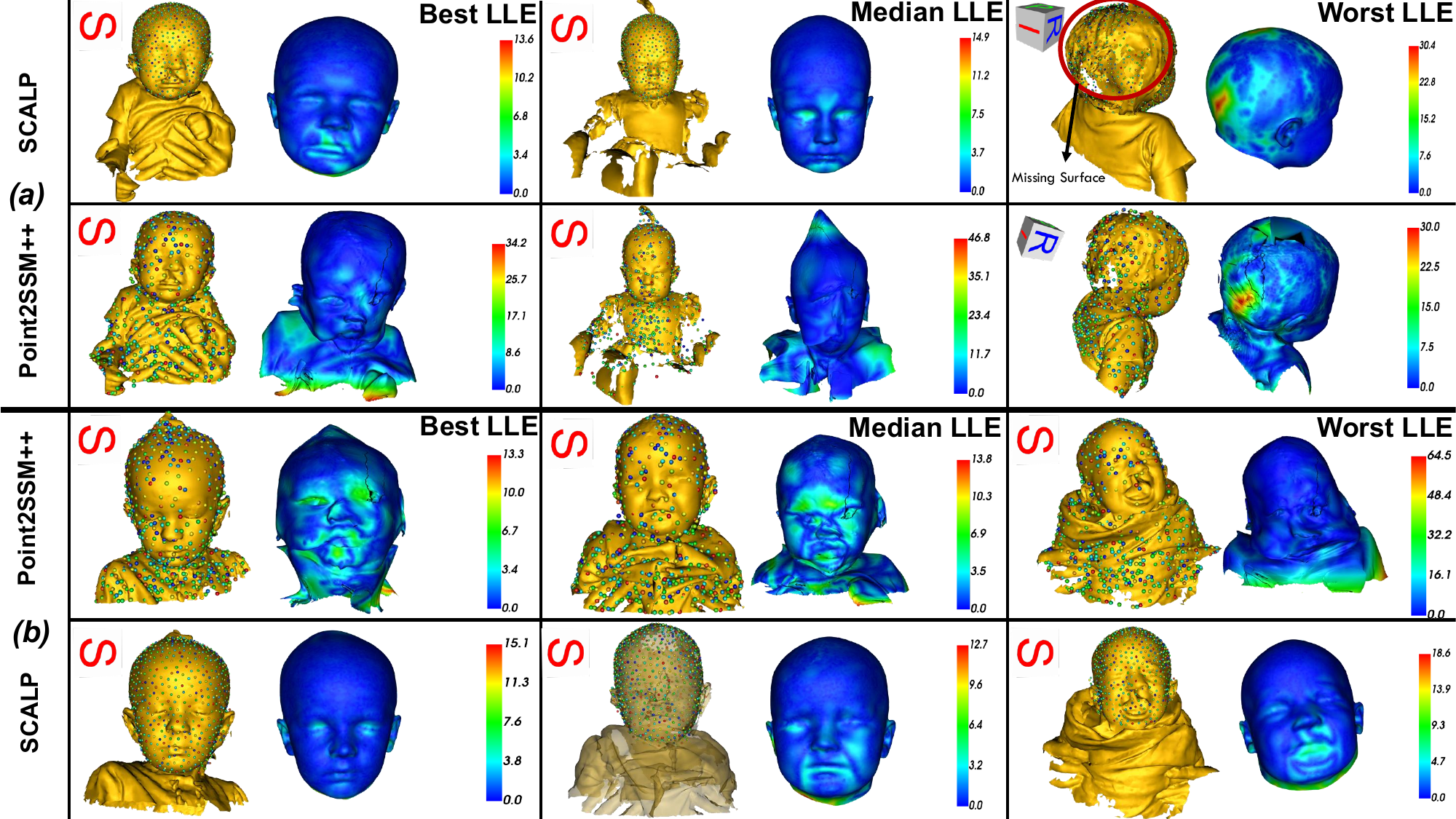}
    \caption{Qualitative comparison between (a) \method{} and (b) Point2SSM++ \cite{point2ssmpp} across best, median, and worst cases (sorted by LLE). Left: predicted particles on raw point clouds. Right: reconstructed meshes colored by local Warp S2S error (mm). Rows pair cross-matched subjects for direct comparison.}
    \label{fig-ssm-qualitative-compare}
  \end{minipage}
  
  \vspace{-20pt}
\end{figure}

\noindent\textit{Quantitative Evaluation and Pipeline Comparison.}
Table~\ref{tab:results_final} and Fig. \ref{fig-ssm-quant} summarizes the quantitative performance of \method{} against the state-of-the-art unsupervised point-cloud correspondence and statistical shape modeling baselines. Across all evaluations, \method{} consistently outperforms the competing methods, often by multiple orders of magnitude. \method{} achieves the lowest Landmark Localization Error (LLE) of $3.520 \pm 0.542$\,mm. Because LLE quantifies the proximity of the established correspondence particles to independently annotated anatomical landmarks, this result demonstrates that our framework recovers a representation that is deeply anchored in clinical anatomy. In contrast, unsupervised baselines, such as Point2SSM++ and DPC, yield significantly higher LLE values ($6.364$\,mm and $5.549$\,mm, respectively). Lacking explicit anatomical constraints, these self-supervised approaches optimize purely for global geometric matching criteria, causing them to miss subtle but clinically vital craniofacial landmarks.
\noindent\textit{Anatomical Isolation vs. Peripheral Clutter.}
The structural failure mode of traditional unsupervised methods becomes glaringly apparent when analyzing the surface reconstruction metrics. Baselines like DG-AE, ISR, and Point2SSM yield highly elevated Point\hyp{}to\hyp{}Surface (P2S) errors, ranging from $13.348$\,mm to $25.896$\,mm. This degradation is a direct consequence of a fundamental capability gap: these methods operate under the implicit assumption that the target point cloud represents an isolated anatomical structure. When confronted with raw, unsegmented clinical photogrammetry scans, their permutation-invariant optimization functions erroneously distribute correspondence particles over non-cranial peripheral clutter, such as the neck, shoulders, and clothing. \method{} entirely bypasses this limitation. 
\noindent\textit{Statistical Shape Model Quality.}
These localized geometric enhancements translate directly into vastly superior population-level shape spaces, as evidenced by the standard SSM metrics (Compactness, Generalization, and Specificity). Unsupervised methods achieve deceptively stable Compactness values (e.g., $0.935$ for DPC and $0.928$ for Point2SSM++), but they fail on Generalization and Specificity, with Chamfer Distances (CD) soaring well above $50$\,mm and $130$\,mm, respectively. This severe divergence occurs because these models are forced to encode highly volatile, non-anatomical noise within their principal components. When generating novel shapes or reconstructing held-out subjects, the linear shape space attempts to extrapolate these unconstrained spatial variations, resulting in unnatural surface distortions and high-frequency folding. On the other hand, \method{} outperforms in all of these metrics.
This is a direct consequence of the mathematical formulation of our pipeline, with band-limited Laplace--Beltrami template warp in Stage II and anchoring landmarks in relevant features. By optimizing, exclusively within the first $r$ low-frequency eigenmodes of the template mesh, our framework filters out high-frequency acquisition noise and extrinsic clutter by construction. The deformation space is physically restricted from generating folded or anatomically implausible surfaces, ensuring that the downstream statistical shape model captures true, smooth, population-level morphological variations of the head. These results demonstrate that \method{} satisfies the stringent geometric and anatomical accuracy requirements necessary to enable automated, scalable, and radiation-free longitudinal monitoring of infant craniosynostosis. Comprehensive ablation studies validating our architectural choices and hyperparameter configurations are provided in Appendix ~\ref{sec:ablations} of the supplementary material.

\subsection{Qualitative Results}
\label{sec:qual}
\noindent\textit{Analysis of Subject-Level Correspondence and Reconstruction.}
To evaluate the clinical reliability of the established surface representations, we correlate our quantitative findings with a visual inspection of individual subject reconstructions. As illustrated in Fig.~\ref{fig-ssm-qualitative-compare}, a stark contrast emerges in how the respective frameworks distribute correspondence particles across raw, unsegmented photogrammetric scans. For the unsupervised baseline Point2SSM++ (Fig.~\ref{fig-ssm-qualitative-compare}b), particles are visibly derailed by non-anatomical structures, erroneously spreading across the neck, shoulders, and arbitrary peripheral noise. Consequently, the resulting mesh reconstructions are highly distorted and fail to preserve the true head geometry of interest. In contrast, \method{} consistently confines its correspondence particles to the true craniofacial anatomy across the best, median, and worst-case subjects. Even in our worst LLE case (Fig.~\ref{fig-ssm-qualitative-compare}a, right), the underlying challenge stems from a fundamental physical limitation of 3D photogrammetry: severe data missingness at the posterior aspect of the cranium caused by camera occlusion or infant hair. While this missing geometry inflates the formal landmark localization error metric, \method{}'s band-limited spectral basis acts as a robust geometric prior. Rather than collapsing or folding inward under data fit pressure, as spatial registration methods do, the template warp maintains structural stability, producing a smooth, uniform particle distribution that preserves high-fidelity morphological alignment with the actual anatomy.
\begin{figure}
\vspace{-20pt}
\centering
\includegraphics[scale=0.2]{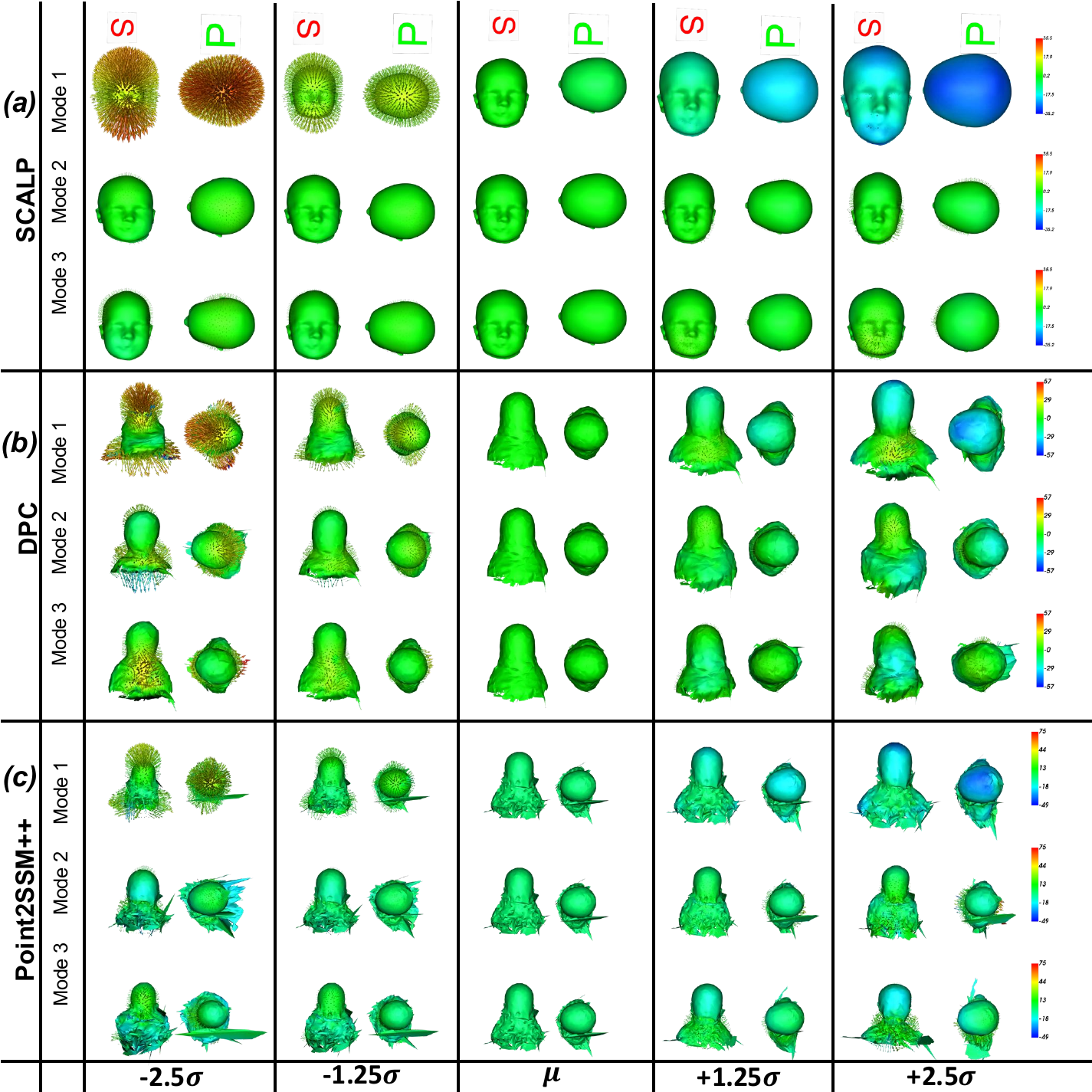}
\caption{First three principal modes of shape variation ($\pm 2.5$ SD from the mean) for (a) \method{}, (b) DPC~\cite{dpc}, and (c) Point2SSM++~\cite{point2ssmpp}. Vectors and heatmaps show deformation direction and signed surface distance relative to the mean shape.}
\label{fig-ssm-qualitative}
\vspace{-20pt}
\end{figure}
\noindent\textit{Interpretation of Population-Level Statistical Shape Spaces.}
The downstream impact of these localized geometric properties is revealed through the principal component analysis (PCA) modes of variation shown in Fig.~\ref{fig-ssm-qualitative}. The statistical shape space generated by \method{} (Fig.~\ref{fig-ssm-qualitative}a) successfully isolates and models clean, clinically relevant anatomical patterns. Specifically, the first principal mode (PC1) captures global head size scaling, primarily accounting for normal variations linked to infant age and sex. The second mode (PC2) accurately captures variations in the Cephalic Index, manifesting as distinct shifts in cranial length and width, which is the primary morphological indicator used to clinically evaluate craniosynostosis phenotypes. The third mode (PC3) isolates mid-facial and structural variations, capturing the distinct phenotypic expressions of the cleft palate subjects included within our diverse cohort. Conversely, the shape spaces generated by the unsupervised baselines, DPC (Fig.~\ref{fig-ssm-qualitative}b) and Point2SSM++ (Fig.~\ref{fig-ssm-qualitative}c), are heavily contaminated. Because these frameworks lack anatomical grounding, their primary modes of variation capture arbitrary combinations of non-head geometry, including shifting neck poses, shoulder orientation, and scanning artifacts. PC1 and PC2 in these baselines encode the scaling of the entire scanning bounding box rather than localized anatomy. By embedding highly volatile peripheral noise into the dominant eigenvectors of the shape model, these unsupervised methods dilute and obscure the true clinical signal. 

\section{Conclusion}
\label{sec:conclusion}

In this work, we presented \method{}, a two-stage hybrid framework designed to construct anatomically grounded and topologically consistent Statistical Shape Models (SSMs) directly from raw, unsegmented infant craniofacial photogrammetry. By decoupling the task into a semi-supervised landmark localization step (Stage I) and a band-limited Laplace--Beltrami spectral deformation engine (Stage II), our pipeline effectively bridges the gap between local clinical anatomy and global population-level geometry. Quantitatively and qualitatively, \method{} demonstrated a profound capability to isolate the infant cranium from peripheral scanning clutter, bypassing the manual preprocessing requirements that cripple standard unsupervised methods. Crucially, our framework achieves state-of-the-art generalizability and specificity while operating under tight data-scarcity constraints, offering a clear path toward objective, automated, and radiation-free longitudinal monitoring of infant craniosynostosis and related craniofacial conditions. \noindent\textit{Limitations and Future Work.}
Despite its strong performance, certain structural limitations remain to be addressed. First, our Stage II deformation engine operates under an implicit mesh modality assumption. If a photogrammetric scan is delivered purely as an unorganized point cloud rather than a surface mesh, an intermediate reconstruction step (e.g., Poisson surface reconstruction) is required before spectral warping. However, this is rarely a bottleneck in practice, as modern clinical stereophotogrammetry systems (e.g., 3dMD) natively output high-quality, calibrated triangulated meshes directly. Second, the pipeline exhibits a dependency on template selection; if the base template mesh deviates morphologically too far from the clinical cohort under consideration, the initialization of the spectral warp can become sensitive to this initial geometric discrepancy. Finally, severe localized surface data missingness at the posterior cranium due to camera occlusion or infant hair can inflate landmark errors, though our band-limited spectral basis prevents drastic mesh collapse in these regions. Future work will focus on validating our established shape descriptors against gold-standard CT-based metrics to evaluate their efficacy for downstream craniosynostosis severity score prediction. Furthermore, because 3D photogrammetry is low-cost and highly accessible, \method{} paves the way toward building large-scale, population-based generative shape models for pediatric craniofacial surgery. Such models could be leveraged to predict post-operative outcomes, simulate surgical interventions, and model longitudinal shape progression conditioned on a specific severity score or phenotypic expression.

\begin{credits}
\subsubsection{\ackname} 
This work was supported by the National Institutes of Health under grant number NIDCR-R01DE032366. The authors thank the CranioRate Consortium, with special appreciation to UPMC Children's Hospital of Pittsburgh for providing the data and clinical expertise used in this study.

\end{credits}
%
%
%
\bibliographystyle{splncs04}

\bibliography{refs}

@article{bhalodia2020,
  title={Quantifying the severity of metopic craniosynostosis: a pilot study application of machine learning in craniofacial surgery},
  author={Bhalodia, Riddhish and Dvoracek, Lucas A and Ayyash, Ali M and Kavan, Ladislav and Whitaker, Ross and Goldstein, Jesse A},
  journal={Journal of Craniofacial Surgery},
  volume={31},
  number={3},
  pages={697--701},
  year={2020},
  publisher={LWW}
}

@inproceedings{deepssm,
  title={DeepSSM: a deep learning framework for statistical shape modeling from raw images},
  author={Bhalodia, Riddhish and Elhabian, Shireen Y and Kavan, Ladislav and Whitaker, Ross T},
  booktitle={International Workshop on Shape in Medical Imaging},
  pages={244--257},
  year={2018},
  organization={Springer}
}

@inproceedings{point2ssm,
  title={Point2SSM: learning morphological variations of anatomies from point clouds},
  author={Adams, Jadie and Elhabian, Shireen},
  booktitle={International Conference on Learning Representations},
  volume={2024},
  pages={13493--13513},
  year={2024}
}

@article{point2ssmpp,
  title={Point2ssm++: Self-supervised learning of anatomical shape models from point clouds},
  author={Adams, Jadie and Karanam, Mokshagna Sai Teja and Elhabian, Shireen},
  journal={Medical Image Analysis},
  pages={104073},
  year={2026},
  publisher={Elsevier}
}

@article{mendoza2014,
  title={Personalized assessment of craniosynostosis via statistical shape modeling},
  author={Mendoza, Carlos S and Safdar, Nabile and Okada, Kazunori and Myers, Emmarie and Rogers, Gary F and Linguraru, Marius George},
  journal={Medical Image Analysis},
  volume={18},
  number={4},
  pages={635--646},
  year={2014},
  publisher={Elsevier}
}

@article{meulstee2017,
  title={A new method for three-dimensional evaluation of the cranial shape and the automatic identification of craniosynostosis using 3D stereophotogrammetry},
  author={Meulstee, JW and Verhamme, LM and Borstlap, WA and Van der Heijden, F and De Jong, GA and Xi, T and Berg{\'e}, SJ and Delye, Hans and Maal, TJJ},
  journal={International journal of oral and maxillofacial surgery},
  volume={46},
  number={7},
  pages={819--826},
  year={2017},
  publisher={Elsevier}
}

@article{rodriguezflorez2017,
  title={Quantifying the effect of corrective surgery for trigonocephaly: a non-invasive, non-ionizing method using three-dimensional handheld scanning and statistical shape modelling},
  author={Rodriguez-Florez, Naiara and G{\"o}ktekin, {\"O}zge K and Bruse, Jan L and Borghi, Alessandro and Angullia, Freida and Knoops, Paul GM and Tenhagen, Maik and O'Hara, Justine L and Koudstaal, Maarten J and Schievano, Silvia and others},
  journal={Journal of Cranio-Maxillofacial Surgery},
  volume={45},
  number={3},
  pages={387--394},
  year={2017},
  publisher={Elsevier}
}

@article{craniorate,
  title={Quantifying sagittal craniosynostosis severity: a machine learning approach with CranioRate},
  author={Tao, Wenzheng and Somorin, Tobi J and Kueper, Janina and Dixon, Angel and Kass, Nicolas and Khan, Nawazish and Iyer, Krithika and Wagoner, Jake and Rogers, Ashley and Whitaker, Ross and others},
  journal={The Cleft Palate Craniofacial Journal},
  pages={10556656251347366},
  year={2025},
  publisher={SAGE Publications Sage CA: Los Angeles, CA}
}

@article{craniosynostosis_overview,
  title={Recent advances in craniosynostosis},
  author={Yilmaz, Elanur and Mihci, Ercan and Nur, Banu and Alper, {\"O}zg{\"u}l M and Ta{\c{c}}oy, {\c{S}}{\"u}kran},
  journal={Pediatric Neurology},
  volume={99},
  pages={7--15},
  year={2019},
  publisher={Elsevier}
}

@article{elkhill2023,
  title={Geometric learning and statistical modeling for surgical outcomes evaluation in craniosynostosis using 3D photogrammetry},
  author={Elkhill, Connor and Liu, Jiawei and Linguraru, Marius George and LeBeau, Scott and Khechoyan, David and French, Brooke and Porras, Antonio R},
  journal={Computer methods and programs in biomedicine},
  volume={240},
  pages={107689},
  year={2023},
  publisher={Elsevier}
}

@inproceedings{cruzguerrero2024,
  title={Mesh Registration via Geometric Feature Homogenization and Offset Cross-Attention: Application to 3D Photogrammetry},
  author={Cruz-Guerrero, In{\'e}s A and Elkhill, Connor and Liu, Jiawei and Nguyen, Phuong and French, Brooke and Porras, Antonio R},
  booktitle={International Workshop on Graphs in Biomedical Image Analysis},
  pages={96--105},
  year={2024},
  organization={Springer}
}

@article{schaufelberger2022,
  title={A radiation-free classification pipeline for craniosynostosis using statistical shape modeling},
  author={Schaufelberger, Matthias and K{\"u}hle, Reinald and Wachter, Andreas and Weichel, Frederic and Hagen, Niclas and Ringwald, Friedemann and Eisenmann, Urs and Hoffmann, J{\"u}rgen and Engel, Michael and Freudlsperger, Christian and others},
  journal={Diagnostics},
  volume={12},
  number={7},
  pages={1516},
  year={2022},
  publisher={MDPI}
}

@article{dejong2020,
  title={Combining deep learning with 3D stereophotogrammetry for craniosynostosis diagnosis},
  author={de Jong, Guido and Bijlsma, Elmar and Meulstee, Jene and Wennen, Myrte and van Lindert, Erik and Maal, Thomas and Aquarius, Ren{\'e} and Delye, Hans},
  journal={Scientific reports},
  volume={10},
  number={1},
  pages={15346},
  year={2020},
  publisher={Nature Publishing Group UK London}
}

@inproceedings{dpc,
  title={Dpc: Unsupervised deep point correspondence via cross and self construction},
  author={Lang, Itai and Ginzburg, Dvir and Avidan, Shai and Raviv, Dan},
  booktitle={2021 International Conference on 3D Vision (3DV)},
  pages={1442--1451},
  year={2021},
  organization={IEEE}
}

@inproceedings{pointnetpp,
  author    = {Qi, Charles R. and Yi, Li and Su, Hao and Guibas, Leonidas J.},
  title     = {{PointNet++}: Deep Hierarchical Feature Learning on Point Sets in
               a Metric Space},
  booktitle = {Advances in Neural Information Processing Systems (NeurIPS)},
  year      = {2017}
}

@article{dgcnn,
  author  = {Wang, Yue and Sun, Yongbin and Liu, Ziwei and Sarma, Sanjay E. and
             Bronstein, Michael M. and Solomon, Justin M.},
  title   = {Dynamic Graph {CNN} for Learning on Point Clouds},
  journal = {ACM Transactions on Graphics},
  volume  = {38},
  number  = {5},
  year    = {2019}
}

@inproceedings{pointtransformer,
  title={Point transformer},
  author={Zhao, Hengshuang and Jiang, Li and Jia, Jiaya and Torr, Philip HS and Koltun, Vladlen},
  booktitle={Proceedings of the IEEE/CVF international conference on computer vision},
  pages={16259--16268},
  year={2021}
}

@article{cpd,
  author  = {Myronenko, Andriy and Song, Xubo},
  title   = {Point Set Registration: Coherent Point Drift},
  journal = {IEEE Transactions on Pattern Analysis and Machine Intelligence},
  volume  = {32},
  number  = {12},
  pages   = {2262--2275},
  year    = {2010}
}

@article{bcpd,
  author  = {Hirose, Osamu},
  title   = {A Bayesian Formulation of Coherent Point Drift},
  journal = {IEEE Transactions on Pattern Analysis and Machine Intelligence},
  volume  = {43},
  number  = {7},
  pages   = {2269--2286},
  year    = {2021}
}

@inproceedings{amberg2007,
  author    = {Amberg, Brian and Romdhani, Sami and Vetter, Thomas},
  title     = {Optimal Step Nonrigid {ICP} Algorithms for Surface Registration},
  booktitle = {IEEE Conference on Computer Vision and Pattern Recognition (CVPR)},
  year      = {2007}
}

@article{ovsjanikov2012,
  author  = {Ovsjanikov, Maks and Ben-Chen, Mirela and Solomon, Justin and
             Butscher, Adrian and Guibas, Leonidas},
  title   = {Functional Maps: A Flexible Representation of Maps Between Shapes},
  journal = {ACM Transactions on Graphics},
  volume  = {31},
  number  = {4},
  year    = {2012}
}

@inproceedings{toothforge,
  author    = {Kub{\'\i}k, Tibor and Guibault, Fran{\c{c}}ois and
               {\v{S}}pan{\v{e}}l, Michal and Lombaert, Herv{\'e}},
  title     = {{ToothForge}: Automatic Dental Shape Generation using
               Synchronized Spectral Embeddings},
  booktitle = {Medical Image Computing and Computer-Assisted Intervention
               (MICCAI)},
  series    = {LNCS},
  publisher = {Springer},
  year      = {2025},
  note      = {arXiv:2506.02702}
}

@inproceedings{isr,
  author    = {Chen, Nenglun and Liu, Lingjie and Cui, Zhiming and Chen, Runnan
               and Ceylan, Duygu and Tu, Changhe and Wang, Wenping},
  title     = {Unsupervised Learning of Intrinsic Structural Representation
               Points},
  booktitle = {IEEE/CVF Conference on Computer Vision and Pattern Recognition
               (CVPR)},
  pages     = {9121--9130},
  year      = {2020}
}

@inproceedings{sc3k,
  author    = {Zohaib, Mohammad and Del Bue, Alessio},
  title     = {{SC3K}: Self-Supervised and Coherent {3D} Keypoints Estimation
               from Rotated, Noisy, and Decimated Point Cloud Data},
  booktitle = {IEEE/CVF International Conference on Computer Vision (ICCV)},
  pages     = {22509--22519},
  year      = {2023}
}

@inproceedings{pointnet,
  author    = {Qi, Charles R. and Su, Hao and Mo, Kaichun and Guibas, Leonidas J.},
  title     = {{PointNet}: Deep Learning on Point Sets for {3D} Classification and
               Segmentation},
  booktitle = {IEEE Conference on Computer Vision and Pattern Recognition (CVPR)},
  year      = {2017}
}

@article{pct,
  author  = {Guo, Meng-Hao and Cai, Jun-Xiong and Liu, Zheng-Ning and Mu,
             Tai-Jiang and Martin, Ralph R. and Hu, Shi-Min},
  title   = {{PCT}: Point Cloud Transformer},
  journal = {Computational Visual Media},
  volume  = {7},
  number  = {2},
  pages   = {187--199},
  year    = {2021}
}

@inproceedings{maturana2015,
  author    = {Maturana, Daniel and Scherer, Sebastian},
  title     = {{VoxNet}: A {3D} Convolutional Neural Network for Real-Time Object
               Recognition},
  booktitle = {IEEE/RSJ International Conference on Intelligent Robots and Systems
               (IROS)},
  year      = {2015}
}

@inproceedings{su2015,
  author    = {Su, Hang and Maji, Subhransu and Kalogerakis, Evangelos and
               Learned-Miller, Erik},
  title     = {Multi-View Convolutional Neural Networks for {3D} Shape Recognition},
  booktitle = {IEEE International Conference on Computer Vision (ICCV)},
  year      = {2015}
}

@inproceedings{masci2015,
  author    = {Masci, Jonathan and Boscaini, Davide and Bronstein, Michael M. and
               Vandergheynst, Pierre},
  title     = {Geodesic Convolutional Neural Networks on Riemannian Manifolds},
  booktitle = {IEEE International Conference on Computer Vision Workshops (ICCVW)},
  year      = {2015}
}

@article{sun2009hks,
  author  = {Sun, Jian and Ovsjanikov, Maks and Guibas, Leonidas},
  title   = {A Concise and Provably Informative Multi-Scale Signature Based on
             Heat Diffusion},
  journal = {Computer Graphics Forum},
  volume  = {28},
  number  = {5},
  pages   = {1383--1392},
  year    = {2009}
}

@inproceedings{aubry2011wks,
  author    = {Aubry, Mathieu and Schlickewei, Ulrich and Cremers, Daniel},
  title     = {The Wave Kernel Signature: A Quantum Mechanical Approach to Shape
               Analysis},
  booktitle = {IEEE International Conference on Computer Vision Workshops (ICCVW)},
  year      = {2011}
}

@article{reuter2006,
  author  = {Reuter, Martin and Wolter, Franz-Erich and Peinecke, Niklas},
  title   = {{Laplace--Beltrami} Spectra as {`Shape-DNA'} of Surfaces and Solids},
  journal = {Computer-Aided Design},
  volume  = {38},
  number  = {4},
  pages   = {342--366},
  year    = {2006}
}

@inproceedings{litany2017,
  title={Deep functional maps: Structured prediction for dense shape correspondence},
  author={Litany, Or and Remez, Tal and Rodola, Emanuele and Bronstein, Alex and Bronstein, Michael},
  booktitle={Proceedings of the IEEE international conference on computer vision},
  pages={5659--5667},
  year={2017}
}

@inproceedings{donati2020,
  title={Deep geometric functional maps: Robust feature learning for shape correspondence},
  author={Donati, Nicolas and Sharma, Abhishek and Ovsjanikov, Maks},
  booktitle={Proceedings of the IEEE/CVF conference on computer vision and pattern recognition},
  pages={8592--8601},
  year={2020}
}

@inproceedings{levy2006,
  title={Laplace-beltrami eigenfunctions towards an algorithm that" understands" geometry},
  author={L{\'e}vy, Bruno},
  booktitle={IEEE International Conference on Shape Modeling and Applications 2006 (SMI'06)},
  pages={13--13},
  year={2006},
  organization={IEEE}
}

@inproceedings{mesh2ssm,
  title={Mesh2ssm: From surface meshes to statistical shape models of anatomy},
  author={Iyer, Krithika and Elhabian, Shireen Y},
  booktitle={International Conference on Medical Image Computing and Computer-Assisted Intervention},
  pages={615--625},
  year={2023},
  organization={Springer}
}

@article{mesh2ssmpp,
  title={Mesh2SSM++: A Probabilistic Framework for Unsupervised Learning of Statistical Shape Model of Anatomies from Surface Meshes},
  author={Iyer, Krithika and Karanam, Mokshagna Sai Teja and Elhabian, Shireen},
  journal={arXiv preprint arXiv:2502.07145},
  year={2025}
}

@article{styner2006,
  author  = {Styner, Martin and Oguz, Ipek and Xu, Shun and Brechb{\"u}hler,
             Christian and Pantazis, Dimitrios and Levitt, James J. and Shenton,
             Martha E. and Gerig, Guido},
  title   = {Framework for the Statistical Shape Analysis of Brain Structures
             Using {SPHARM-PDM}},
  journal = {Insight Journal},
  pages   = {242--250},
  year    = {2006}
}

@article{tarvainen2017meanteacher,
  title={Mean teachers are better role models: Weight-averaged consistency targets improve semi-supervised deep learning results},
  author={Tarvainen, Antti and Valpola, Harri},
  journal={Advances in neural information processing systems},
  volume={30},
  year={2017}
}

@article{ptv2,
  title={Point transformer v2: Grouped vector attention and partition-based pooling},
  author={Wu, Xiaoyang and Lao, Yixing and Jiang, Li and Liu, Xihui and Zhao, Hengshuang},
  journal={Advances in Neural Information Processing Systems},
  volume={35},
  pages={33330--33342},
  year={2022}
}

@inproceedings{qi2019votenet,
  title={Deep hough voting for 3d object detection in point clouds},
  author={Qi, Charles R and Litany, Or and He, Kaiming and Guibas, Leonidas J},
  booktitle={proceedings of the IEEE/CVF International Conference on Computer Vision},
  pages={9277--9286},
  year={2019}
}

@article{khan2023statistical,
  title={Statistical multi-level shape models for scalable modeling of multi-organ anatomies},
  author={Khan, Nawazish and Peterson, Andrew C and Aubert, Benjamin and Morris, Alan and Atkins, Penny R and Lenz, Amy L and Anderson, Andrew E and Elhabian, Shireen Y},
  journal={Frontiers in Bioengineering and Biotechnology},
  volume={11},
  pages={1089113},
  year={2023},
  publisher={Frontiers Media SA}
}

@incollection{shapeworks,
  title={Shapeworks: particle-based shape correspondence and visualization software},
  author={Cates, Joshua and Elhabian, Shireen and Whitaker, Ross},
  booktitle={Statistical shape and deformation analysis},
  pages={257--298},
  year={2017},
  publisher={Elsevier}
}

@article{davies2002minimum,
  title={A minimum description length approach to statistical shape modeling},
  author={Davies, Rhodri H and Twining, Carole J and Cootes, Timothy F and Waterton, John C and Taylor, Christopher J},
  journal={IEEE transactions on medical imaging},
  volume={21},
  number={5},
  pages={525--537},
  year={2002},
  publisher={IEEE}
}

@article{zhang2022point,
  title={Point-m2ae: multi-scale masked autoencoders for hierarchical point cloud pre-training},
  author={Zhang, Renrui and Guo, Ziyu and Gao, Peng and Fang, Rongyao and Zhao, Bin and Wang, Dong and Qiao, Yu and Li, Hongsheng},
  journal={Advances in neural information processing systems},
  volume={35},
  pages={27061--27074},
  year={2022}
}

@inproceedings{achlioptas2018learning,
  title={Learning representations and generative models for 3d point clouds},
  author={Achlioptas, Panos and Diamanti, Olga and Mitliagkas, Ioannis and Guibas, Leonidas},
  booktitle={International conference on machine learning},
  pages={40--49},
  year={2018},
  organization={PMLR}
}

@article{abdel2023sagittal,
  title={Sagittal craniosynostosis: comparing surgical techniques using 3D photogrammetry},
  author={Abdel-Alim, Tareq and Kurniawan, Melissa and Mathijssen, Irene and Dremmen, Marjolein and Dirven, Clemens and Niessen, Wiro and Roshchupkin, Gennady and van Veelen, Marie-Lise},
  journal={Plastic and Reconstructive Surgery},
  volume={152},
  number={4},
  pages={675e--688e},
  year={2023},
  publisher={LWW}
}

@article{abdel2023reliability,
  title={Reliability and agreement of automated head measurements from 3-dimensional photogrammetry in young children},
  author={Abdel-Alim, Tareq and Tio, Pauline and Kurniawan, Melissa and Mathijssen, Irene and Dirven, Clemens and Niessen, Wiro and Roshchupkin, Gennady and van Veelen, Marie-Lise},
  journal={Journal of Craniofacial Surgery},
  volume={34},
  number={6},
  pages={1629--1634},
  year={2023},
  publisher={LWW}
}

@article{bruce20233d,
  title={3D photography to quantify the severity of metopic craniosynostosis},
  author={Bruce, Madeleine K and Tao, Wenzheng and Beiriger, Justin and Christensen, Cameron and Pfaff, Miles J and Whitaker, Ross and Goldstein, Jesse A},
  journal={The Cleft Palate Craniofacial Journal},
  volume={60},
  number={8},
  pages={971--979},
  year={2023},
  publisher={SAGE Publications Sage CA: Los Angeles, CA}
}

@inproceedings{dai2018non,
  title={Non-rigid 3D shape registration using an adaptive template},
  author={Dai, Hang and Pears, Nick and Smith, William},
  booktitle={Proceedings of the European Conference on Computer Vision (ECCV) Workshops},
  pages={0--0},
  year={2018}
}

@inproceedings{rodola2017partial,
  title={Partial functional correspondence},
  author={Rodol{\`a}, Emanuele and Cosmo, Luca and Bronstein, Michael M and Torsello, Andrea and Cremers, Daniel},
  booktitle={Computer graphics forum},
  volume={36},
  number={1},
  pages={222--236},
  year={2017},
  organization={Wiley Online Library}
}

@misc{abdelalim2023craniumpy,
  author    = {Abdel-Alim, Tarek},
  title     = {CraniumPy: A tool for the alignment and basic analysis of 3D meshes, focused on craniofacial applications},
  year      = {2021},
  publisher = {GitHub},
  journal   = {GitHub Repository},
  howpublished = {\url{https://github.com/T-AbdelAlim/CraniumPy}},
  doi       = {10.5281/zenodo.5634153}
}

\newpage
\appendix
\section*{Supplementary Material}
\renewcommand{\thesection}{\Alph{section}}
\setcounter{section}{0}

\section{Ablation Studies}
\label{sec:ablations}

We conduct five ablation studies to isolate and validate the individual architectural contributions of our two-stage framework.

\noindent\textbf{Semi-Supervised vs. Fully Supervised Learning (Stage I):}
To isolate our semi-supervised recipe, we train the Stage I landmark detector strictly on the 50 labeled subjects, completely disabling the Mean Teacher consistency objective and pseudo-labeling. The resulting landmarks are passed through an identical Stage II engine. As shown in Table~\ref{tab:ablation_main}, while local surface fit metrics (P2S, Warp S2S) remain comparable, the supervised baseline exhibits a catastrophic degradation in downstream SSM metrics: Generalization error more than triples ($3.402$\,mm vs.\ $1.112$\,mm) and Specificity significantly worsens ($16.310$\,mm vs.\ $9.604$\,mm). Furthermore, the fully supervised model displays a $4\times$ increase in LLE variance ($\pm 2.005$\,mm vs.\ $\pm 0.542$\,mm), indicating severe overfitting. This landmark jitter disrupts topological consistency across the population, forcing the linear PCA space to encode alignment noise rather than true morphological anatomy.

\noindent\textbf{Spectral vs. Spatial Deformation (Stage II):}
To isolate our Stage II Laplace--Beltrami spectral framework, we replace it with a standard spatial baseline: Thin Plate Spline (TPS) deformation using the identical Stage I predicted landmarks as control points. As presented in Table~\ref{tab:ablation_main}, full \method{} outperforms the Stage I + TPS pipeline across every surface and statistical metric. Notably, the TPS baseline exhibits a major degradation in Specificity ($19.034$\,mm vs.\ $9.604$\,mm) and nearly triples the P2S reconstruction error ($4.340$\,mm vs.\ $1.478$\,mm). When confronted with large missing surface regions due to hair or camera occlusion, TPS extrapolates unconstrained deformations outside the sparse landmark set, causing high-frequency wrinkling. Conversely, \method{}'s band-limited eigenbasis imposes a hard geometric constraint that prevents high-frequency folding by construction.

\begin{table}[htbp]
\vspace{-30pt}
\centering
\small
\setlength{\tabcolsep}{4pt}
\renewcommand{\arraystretch}{0.85}
\caption{Ablation analyses isolating Stage I (Semi-supervised vs.\ Fully Supervised) and Stage II (Spectral vs.\ TPS Spatial warping).}
\label{tab:ablation_main}
\begin{tabular}{lrrrr}
\toprule
\textbf{Metric} & \textbf{\method{}} & \textbf{Stage I Sup.} & \textbf{Stage I + TPS} \\
\midrule
P2S (mm) $\downarrow$ & $\mathbf{1.478 \pm 0.497}$ & $1.492 \pm 0.604$ & $4.340 \pm 1.552$ \\
Warp S2S (mm) $\downarrow$ & $\mathbf{1.456 \pm 0.492}$ & $1.469 \pm 0.583$ & $4.270 \pm 1.531$ \\
LLE (mm) $\downarrow$ & $\mathbf{3.520 \pm 0.542}$ & $3.929 \pm 2.005$ & $4.644 \pm 1.113$ \\
Compactness $\uparrow$ & $\mathbf{0.990}$ & $0.984$ & $0.986$ \\
Generalization (CD) $\downarrow$ & $\mathbf{1.112}$ & $3.402$ & $1.487$ \\
Specificity (CD) $\downarrow$ & $\mathbf{9.604}$ & $16.310$ & $19.034$ \\
\bottomrule
\end{tabular}
\vspace{-20pt}
\end{table}

\noindent\textbf{Upstream Prediction vs. Oracle Landmarks:}
To decouple Stage I detector errors from the Stage II warping engine, we evaluate an ``Oracle'' configuration by replacing predicted landmarks on the 13 held-out test subjects with expert manual annotations. Note that because a population-wide SSM requires computing a PCA basis over the entire training set (where manual oracle annotations are intentionally omitted), global SSM metrics cannot be computed for this localized test ablation. As shown in Table~\ref{tab:ablation_oracle}, transitioning to manual landmarks yields only a modest, incremental improvement (e.g., Warp S2S error decreases by merely $\approx 0.36$\,mm). This narrow performance margin demonstrates that our upstream Stage I detector has already achieved a high accuracy ceiling and is not a pipeline bottleneck, while confirming that Stage II successfully accommodates minor localized sub-millimeter landmark variances.

\begin{table}[htbp]
\vspace{-30pt}
\centering
\small
\setlength{\tabcolsep}{6pt}
\renewcommand{\arraystretch}{0.85}
\caption{Ablation analysis decoupling Stage I and Stage II error budgets on test subjects using manual expert Oracle Landmarks.}
\label{tab:ablation_oracle}
\begin{tabular}{lrr}
\toprule
\textbf{Metric} & \textbf{\method{} (Predicted)} & \textbf{Oracle Landmarks + Stage II} \\
\midrule
P2S (mm) $\downarrow$ & $1.478 \pm 0.497$ & $\mathbf{1.106 \pm 0.406}$ \\
Warp S2S (mm) $\downarrow$ & $1.456 \pm 0.492$ & $\mathbf{1.096 \pm 0.410}$ \\
LLE (mm) $\downarrow$ & $3.520 \pm 0.542$ & $\mathbf{3.203 \pm 0.494}$ \\
\bottomrule
\end{tabular}
\vspace{-10pt}
\end{table}

\begin{figure}[htbp]
\vspace{-23pt}
  \centering
  \begin{minipage}[b]{0.43\textwidth}
    \centering
    \includegraphics[width=\textwidth]{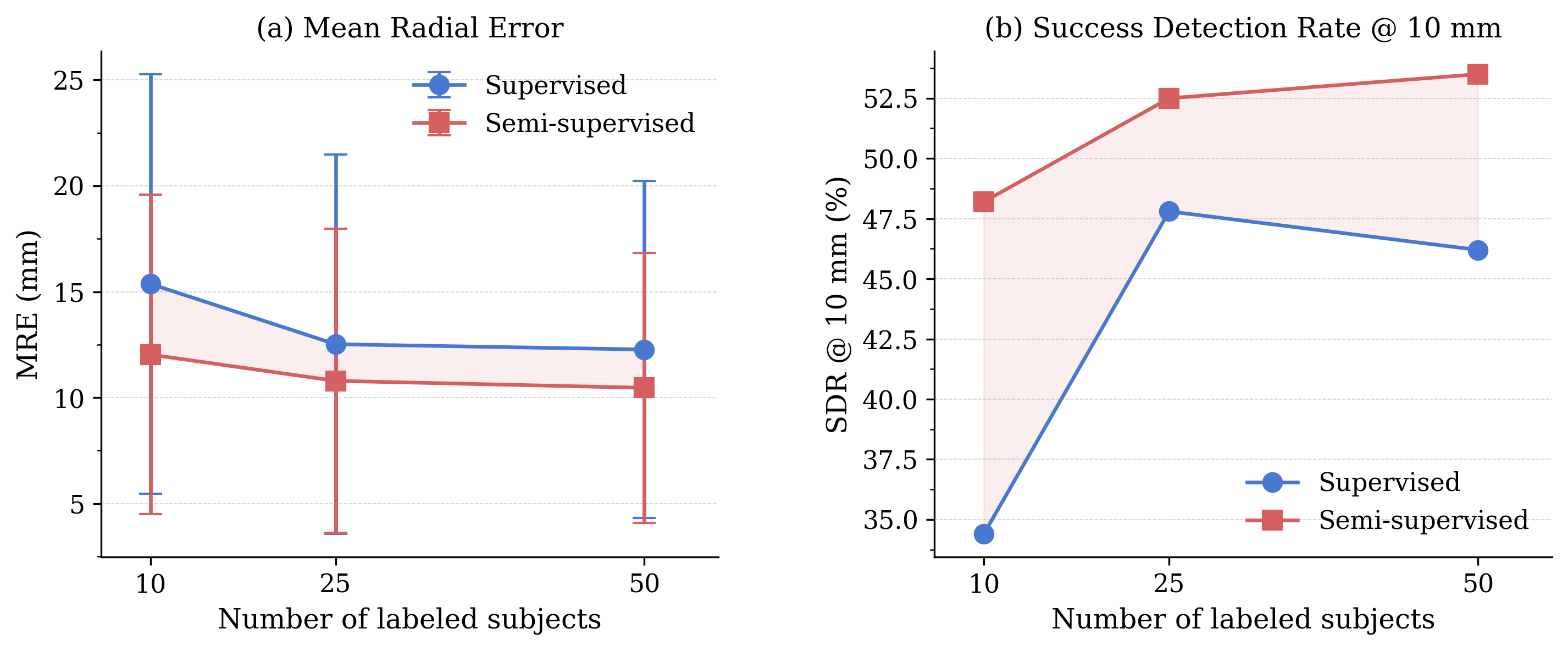}
    \caption{Data efficiency ablation curves on the test cohort for varying quantities of labeled training subjects ($N \in \{10, 25, 50\}$).
    }
    \label{fig-data-efficiency}
  \end{minipage}
  \hfill 
  \begin{minipage}[b]{0.55\textwidth}
    \centering
    \setlength{\tabcolsep}{1.5pt} 
    \renewcommand{\arraystretch}{0.85}
    \captionof{table}{Ablation analysis on the LBO spectral mode truncation rank $r$.}
    \label{tab:ablation_rank}
    \tiny
    \resizebox{\textwidth}{!}{%
      \begin{tabular}{lrrrrr}
      \toprule
      \textbf{Metric} & \textbf{$r=64$} & \textbf{$r=100$} & \textbf{\method{} ($r=120$)} & \textbf{$r=150$} & \textbf{$r=200$} \\
      \midrule
      P2S (mm) $\downarrow$ & $1.150 \pm 0.408$ & $1.102 \pm 0.394$ & $1.478 \pm 0.497$ & $1.081 \pm 0.386$ & $\mathbf{1.062 \pm 0.376}$ \\
      Warp S2S $\downarrow$ & $1.134 \pm 0.406$ & $1.089 \pm 0.393$ & $1.456 \pm 0.492$ & $1.068 \pm 0.385$ & $\mathbf{1.049 \pm 0.379}$ \\
      LLE (mm) $\downarrow$ & $3.469 \pm 0.505$ & $3.416 \pm 0.518$ & $3.520 \pm 0.542$ & $\mathbf{3.380 \pm 0.496}$ & $3.384 \pm 0.508$ \\
      Compactness $\uparrow$ & $0.985$ & $0.985$ & $\mathbf{0.990}$ & $0.985$ & $0.985$ \\
      Generalization $\downarrow$ & $2.758$ & $2.982$ & $\mathbf{1.112}$ & $2.969$ & $3.041$ \\
      Specificity $\downarrow$ & $12.163$ & $11.864$ & $\mathbf{9.604}$ & $12.254$ & $12.295$ \\
      \bottomrule
      \end{tabular}%
    }
    \vspace{10pt}
  \end{minipage}
  \vspace{-10pt}
\end{figure}

\noindent\textbf{Tuning of Spectral Basis Dimensionality $r$:}
The truncation rank $r$ of the LBO eigenfunctions dictates the deformation space's expressiveness. We evaluate our pipeline across five capacities: $r \in \{64, 100, 120, 150, 200\}$. As presented in Table~\ref{tab:ablation_rank}, increasing $r$ from 64 to 200 steadily reduces individual surface fitting errors (P2S and Warp S2S). However, optimizing exclusively for surface alignment causes a severe regression in population-level SSM utility. Beyond $r=120$, Generalization and Specificity metrics deteriorate significantly (e.g., Generalization increases from $1.112$\,mm at $r=120$ to $3.041$\,mm at $r=200$). This structural behavior confirms that excessive high-frequency spectral modes enable the model to encode localized acquisition noise into the correspondences, corrupting the downstream linear PCA shape space. Selecting $r=120$ establishes the optimal equilibrium.

\noindent\textbf{Annotation Data Efficiency:}
We evaluate framework resilience under sparse data constraints by varying the labeled training size ($N \in \{10, 25, 50\}$; Fig.~\ref{fig-data-efficiency}). While the supervised baseline degrades steeply—its Success Detection Rate (SDR@10) plummeting to $34.4\%$ at $N=10$ with high variance, our semi-supervised approach degrades gracefully. At $N=10$, \method{} achieves an SDR@10 of $48.2\%$, outperforming the supervised baseline trained on over double the data ($N=25$) while narrowing error variance. This demonstrates that consistency regularization effectively leverages unlabeled cohorts to minimize manual annotation bottlenecks.

\end{document}